\documentclass{bmvc2k}

\title{Deformation-Guided Unsupervised Non-Rigid Shape Matching}
\addauthor{Aymen Merrouche}{aymen.merrouche@inria.fr}{1}
\addauthor{Jo\~ao Regateiro}{joaoregateiro@gmail.com}{2}
\addauthor{Stefanie Wuhrer}{stefanie.wuhrer@inria.fr}{1}
\addauthor{Edmond Boyer}{edmond.boyer@inria.fr}{1}

\addinstitution{
Inria centre\\University Grenoble Alpes\\Grenoble, France
}
\addinstitution{
 Interdigital\\ Cesson-Sévigné, France
}

\runninghead{Merrouche ET AL}{Deformation-Guided Unsupervised Shape Matching}

\def\eg{\emph{e.g}\bmvaOneDot}

\def\etal{\emph{et al}\bmvaOneDot}

\usepackage[bottom]{footmisc}
\usepackage{makecell}
\usepackage{multirow}
\usepackage{wrapfig}
\usepackage{float}
\usepackage{mathbbol}
\usepackage{comment}
\usepackage{pifont}
\newcommand{\cmark}{\ding{51}}%
\newcommand{\xmark}{\ding{55}}%

\usepackage{diagbox}

\begin{document}

\maketitle

\begin{abstract}
We present an unsupervised data-driven approach for non-rigid shape matching. Shape matching identifies correspondences between two shapes and is a fundamental step in many computer vision and graphics applications. Our approach is designed to be particularly robust when matching shapes digitized using 3D scanners that contain fine geometric detail and suffer from different types of noise including topological noise caused by the coalescence of spatially close surface regions. We build on two strategies. First, using a hierarchical patch based shape representation we match shapes consistently in a coarse to fine manner, allowing for robustness to noise. This multi-scale representation drastically reduces the dimensionality of the problem when matching at the coarsest scale, rendering unsupervised learning feasible. Second, we constrain this hierarchical matching to be reflected in 3D by fitting a patch-wise near-rigid deformation model. Using this constraint, we leverage spatial continuity at different scales to capture global shape properties, resulting in matchings that generalize well to data with different deformations and noise characteristics. Experiments demonstrate that our approach obtains significantly better results on raw 3D scans than state-of-the-art methods, while performing on-par on standard test scenarios. Our code is publicly available at \href{https://gitlab.inria.fr/amerrouc/deformation-guided-unsupervised-non-rigid-shape-matching}{https://gitlab.inria.fr/amerrouc/deformation-guided-unsupervised-non-rigid-shape-matching}.
\end{abstract}



\section{Introduction}
\label{sec:introduction}

\begin{figure}[t]
\centering
  \includegraphics[width=0.7\textwidth]{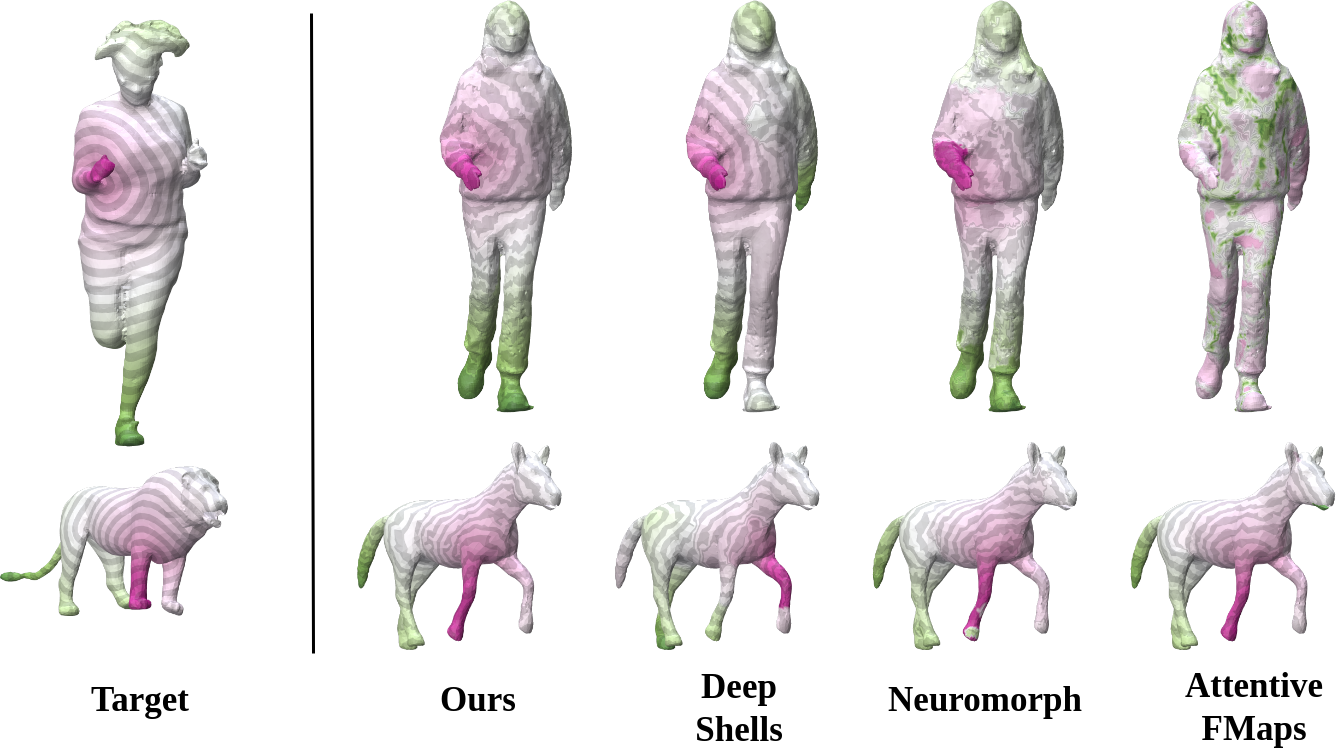}
  \caption{Overview of input, output and our results compared to Deep Shells \cite{eisenberger2020deep}, Neuromorph \cite{eisenberger2021neuromorph} and AttentiveFMaps \cite{li2022attentivefmaps} on challenging data. Each point on target (left) is assigned a color, which is transferred to the source (right) using correspondences computed by different methods. Our results are both globally and locally correct.}
  \label{fig:qualitataive_eval}
\end{figure}

Non-rigid shape matching is the problem of finding correspondences between two geometric objects which differ in their pose or shape. It provides deformation or transfer functions between shapes, including motion fields when considering a single object in motion. Shape matching is a critical part of many applications in computer vision and computer graphics. Recently, data-driven approaches that learn plausible correspondences from training datasets have shown significant improvements in matching shapes with complex deformations, for instance humans with clothing. Yet, they often require ground truth correspondences for training. We consider the unsupervised case where no ground truth is provided.

Shape matching  has been studied extensively over the past decades. Recent data-driven strategies include matching shapes in a spectral basis or adversely in the spatial domain. Spectral methods~\eg~\cite{litany2017deep,li2022attentivefmaps} find global maps between real-valued functions defined in spectral shape spaces, allowing them to generalize to different deformations and sampling rates. However, correct shape topology is assumed for spectral decompositions. Spatial methods learn the alignment of shapes~\eg~\cite{groueix20183d,lang2021dpc,zeng2021corrnet3d} or consider the registration to a given template by casting the problem as a vertex classification~\eg~\cite{monti2017geometric, fey2018splinecnn}. They are robust to changes in topology, but exhibit low generalisation abilities to deformations unseen during training due to the unstructured feature space used for matching. 

This demonstrates that matching two shapes robustly remains difficult, especially when considering raw 3D scans. In addition to large non-rigid deformations and different sampling distributions of the shapes, raw 3D scans suffer from sensor noises including geometric noise of the vertex positions, and topological noise caused by the coalescence of spatially close surface regions. These noises distort both the extrinsic and intrinsic geometry of the shape that is being captured. Matching raw 3D scans remains a challenging problem.

We design a robust matching technique that generalizes well and is robust to noise including topological noise present in raw 3D scans. This is achieved by a spatial matching method that retains the robustness of spatial approaches while producing correspondence maps that, as in the case of spectral approaches, capture global geometric shape properties. 

This is done by relying on two strategies. First, our method considers a hierarchical approach that builds correspondences at different scales of the shapes. We associate shape elements in a coarse-to-fine approach going from coarse surface patches to vertices, which has two main advantages. First, it significantly increases robustness to noise. Second, it reduces the dimensionality of the problem, which allows for efficient unsupervised learning. At coarse scales, we can represent the matching between two shapes as a small matrix, on which we can efficiently impose desirable properties such as cycle consistency or minimal distortion. This information can subsequently be leveraged efficiently at finer scales. 

Second, our method enforces correspondences to agree with a deformation model that controls the spatial continuity of the produced matching. We choose a piece-wise near-rigid deformation model that can represent complex non-rigid deformations, and successfully models deformations of humans, possibly in clothing, and other vertebrates. Imposing spatial continuity on our hierarchical maps constrains correspondences as a whole by linking the matching of individual shape elements, thus allowing to capture global shape properties.

To combine these two strategies, an association network estimating the matching between two shapes is combined with a deformation network estimating the induced alignment in 3D. Both networks operate in a hierarchical fashion, which allows for unsupervised training. Although
multi-scale matching and deformation guidance were used in optimisation based matching approaches~\eg~\cite{cagniart, sipiran2013fully, bonarrigo2014deformable, cao2015two}, the novelty of our work is the design of a data-driven method that combines these two strategies in an unsupervised fashion to build a structured feature space where shapes can be embedded and matched with robustness to noise and generalisation to different deformations and sampling rates.

We demonstrate experimentally that our method performs on-par with state-of-the-art for standard test scenarios. Furthermore, our approach significantly outperforms existing methods when matching raw 3D scans after having been trained exclusively on clean data. Figure~\ref{fig:qualitataive_eval} shows this for challenging data with topological noise (top row). 

In summary, our main contributions are as follows. Firstly, we propose a novel spatial unsupervised data-driven non-rigid shape matching approach that combines multi-scale association maps with a piecewise near-rigid deformation model. Secondly, we outperform state-of-the-art on matching raw 3D scans captured using a multi-camera platform.

\section{Related Work}
\label{sec:related_work}
Non-rigid shape matching has been studied extensively during the past decades. We focus on unsupervised learning-based approaches as our method falls into this category. For a deeper review we refer to a recent survey \cite{deng2022survey}. Existing works can be roughly categorised into two main classes : spatial methods and spectral methods.

\textbf{Spatial Methods}
Early spatial methods~\cite{groueix20183d, deprelle2019learning} proposed supervised strategies to extract a matching between point clouds by learning the deformation in 3D space that best aligns the shapes to a common template. Groueix~\etal~\cite{groueix2019unsupervised} use an unsupervised network for point cloud matching by template-free deformation, where cycle consistency of the deformation is used as a supervising signal to define good correspondences. Recent methods embed input shapes in a feature space where the shapes are aligned guided by criteria on the alignment in 3D. CorrNet3D~\cite{zeng2021corrnet3d} and DPC~\cite{lang2021dpc} estimate an association in feature space between two point clouds. CorrNet3D guides this association by the reconstruction of the two point clouds, while DPC uses self- and cross-reconstruction of the two point clouds for this purpose. Closely related in spirit to our method is NeuroMorph~\cite{eisenberger2021neuromorph}, an unsupervised mesh matching method that simultaneously learns associations in feature space and an interpolation between two meshes in 3D. The interpolation is used as a criterion to guide the correspondence search in feature space. In our method, the deformation search is used to guide the global correspondence search in addition to the hierarchical modeling of the matching.

\textbf{Spectral Methods}
Functional maps (FM)~\cite{ovsjanikov2012functional} introduced the idea of considering point-to-point matching as a special case of mappings between functions defined on shapes, and forms the basis of many matching algorithms. Functions defined on points are projected onto the eigenfunctions of the Laplace-Beltrami operator of the shapes, where the mapping between functions is estimated. This mapping is then used to extract a point-to-point matching. Deep functional maps were introduced in Litany~\etal~\cite{litany2017deep} where instead of using pre-defined point descriptors as the functions to match, the output of a neural network is used. Deep functional maps were successfully extended to the unsupervised regime, by minimizing a distortion measure of the extracted point map~\cite{halimi2019unsupervised, ginzburg2020cyclic}, or by exploiting the desired structural properties of the maps directly in the spectral domain~\cite{roufosse2019unsupervised, sharma2020weakly}. DUO-FM~\cite{donati2022deep} proposes to incorporate the complex functional maps in the representation to estimate orientation preserving maps. More recently, AttentiveFMaps~\cite{li2022attentivefmaps} propose a spectral attention framework to combine multiple resolution functional maps producing maps that are particularly robust at handling non-isometric shapes. 

Other spectral approaches combine the alignment in a spectral shape space with an alignment in ambient space. DeepShells~\cite{eisenberger2020deep} is a functional map based method that also aligns shapes in 3D. The learning criterion of the map measures the alignment tightness between the deformed source and the target shape in ambient space. Spectral Teacher Spatial Student (STS)~\cite{efroni2022spectral} proposes a student teacher mechanism where one of the spatial methods DPC~\cite{lang2021dpc} or CorrNet3D~\cite{zeng2021corrnet3d} is used as a student network to embed shapes in a feature space where the point-to-point mapping is estimated via feature similarity. These features are then used as functions to match in a teacher standard functional maps mechanism. STS uses the spectral teacher to build a feature space that captures global shape properties. We use the deformation model constraint at multiple scales to achieve the same goal.

\section{Method}
\label{sec:method}

This section details our unsupervised\footnote{We assume training shapes without correspondence information that are rigidly pre-aligned in $\mathbb{R}^3$ to have roughly the same front-back orientation. This is sometimes referred to as weakly supervised in the literature.} data-driven matching approach. Given as input two 3D meshes $\mathcal{X}$ and $\mathcal{Y}$, our method outputs a hierarchical mapping between $\mathcal{X}$ and $\mathcal{Y}$ as well as the deformation in 3D that this mapping induces.

Our method represents $\mathcal{X}$ and $\mathcal{Y}$ hierarchically with a hierarchy of surface patches using a greedy approach based on furthest point sampling inspired by \cite{peyre2006geodesic}. We select $L+1$ patch resolutions where patch $0$ represents the vertex level (where each patch is restricted to the vertex itself) and $L$ represents the coarsest patch level. In the following we denote the patches as $(P^l_i)_{1 \leq i \leq n_l}$ for $l=0,\ldots,L$ and their centers as $\mathcal{C}_l = (c^l_i \in \mathbb{R}^3)_{1 \leq i \leq n_l}$.

\begin{figure}
	\centering
	\includegraphics[width=0.9\linewidth]{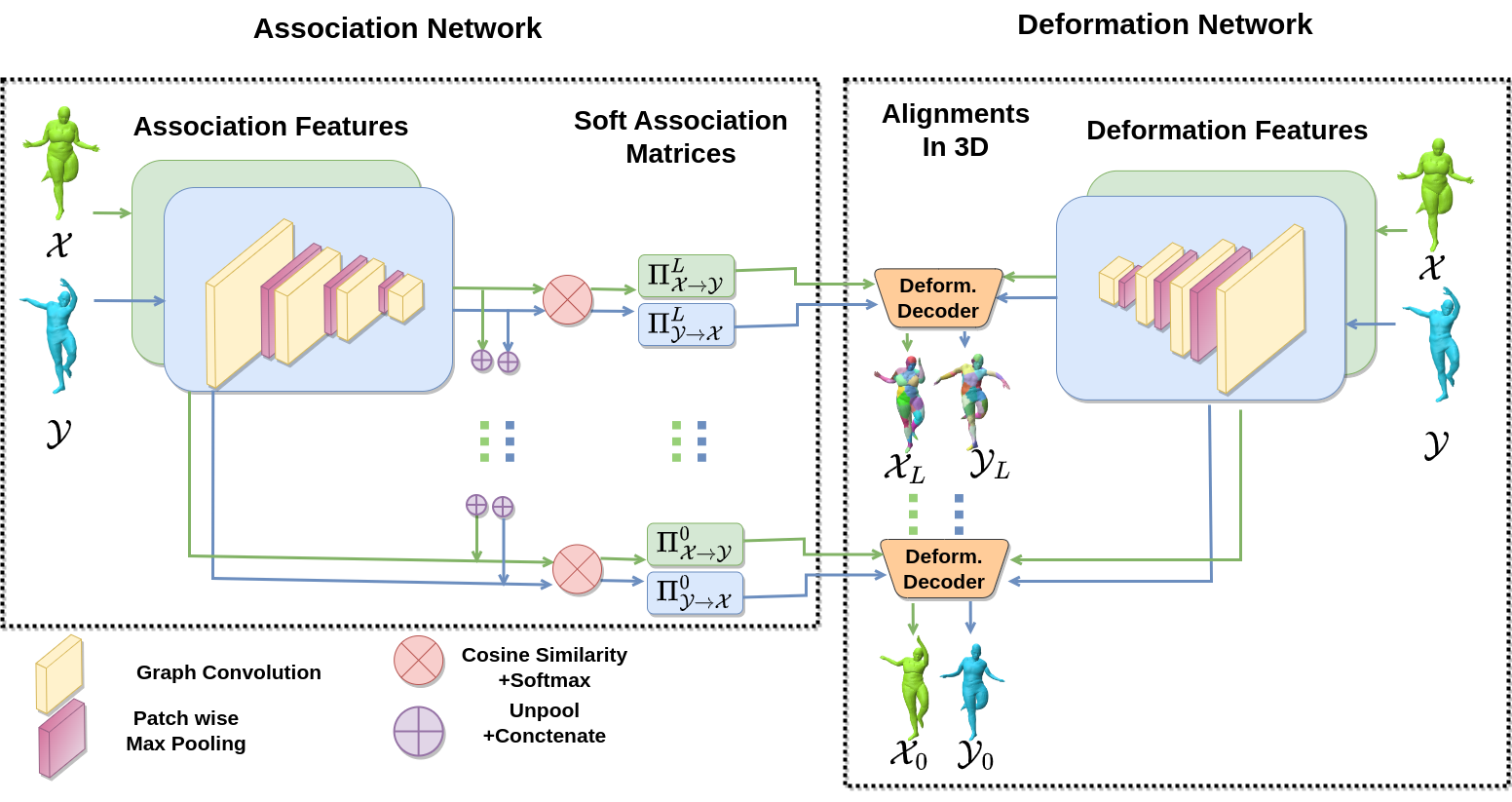}
	\caption{Network architecture. The network takes as input $\mathcal{X}, \mathcal{Y}$ decomposed into a hierarchy of surface patches. It is decomposed into two networks that work from coarse to fine levels, sequentially per level. An association network extracts coarse-to-fine correspondences between $\mathcal{X}$ and $\mathcal{Y}$ as inter-patch association matrices (from $\Pi^L$ to $\Pi^0$). A deformation network outputs deformations that respect these associations in 3D (from $\mathcal{X}_L$, $\mathcal{Y}_L$ to $\mathcal{X}_0$, $\mathcal{Y}_0$).}
	\label{fig:network_architecture}
\end{figure}

Our method is implemented by a composition of two networks. First, an association network that aligns $\mathcal{X}$ and $\mathcal{Y}$ in a hierarchical feature space as detailed in Sec.~\ref{sec:association_network}. Second, a deformation network that constrains these hierarchical maps by aligning $\mathcal{X}$ and $\mathcal{Y}$ in 3D space by fitting a hierarchical patch-wise near-rigid deformation model as detailed in Sec.~\ref{sec:deformation_network}. Fig.~\ref{fig:network_architecture} shows the architecture of our network.

\subsection{Association Network}
\label{sec:association_network}
This network estimates association maps hierarchically, from coarse patches to vertices. Association maps are computed as matrices $\Pi^l_{\mathcal{X} \rightarrow \mathcal{Y}}$ and $\Pi^l_{\mathcal{Y} \rightarrow \mathcal{X}}$ mapping from surface patches of $\mathcal{X}$ to surface patches of $\mathcal{Y}$ and vice versa. Element $(i,j)$ of $\Pi^l_{\mathcal{X} \rightarrow \mathcal{Y}}$ contains a matching score between the $i$-th patch of $\mathcal{X}$ and the $j$-th patch of $\mathcal{Y}$ at hierarchy level $l$, estimated via feature similarity. The shape features are extracted by a convolutional graph neural network, based on the FeaStConv~\cite{verma2018feastnet} operator. At hierarchy level $l$, FeaStConv acts on patch neighborhoods. On the vertex scale (hierarchy level $0$), where convolutions act on vertex neighborhoods, we use vertex coordinates and normals as input features. We use patch-wise max-pooling to go to a coarser hierarchy level. This allows to compute patch-wise features for hierarchy level $l$ on $\mathcal{X}$ and $\mathcal{Y}$, which we denote by $\mathbb{X}^{l} = (\mathbb{x}^l_i \in \mathbb{R}^{d_l} )_{1 \leq i \leq n_l}$ and $\mathbb{Y}^{l} = (\mathbb{y}^l_i \in \mathbb{R}^{d_l})_{1 \leq i \leq m_l}$, where $n_l$, $m_l$ are the number of patches at level $l$ for $\mathcal{X}$ and $\mathcal{Y}$. 

Features are computed in a fine-to-coarse manner. To allow for coarse-to-fine hierarchical matching, we combine the feature of a patch at level $l$ with the features of all its parent patches in coarser levels $l+1,\ldots,L$. We denote these combined features by $\mathbb{X}^{L, l} = (\mathbb{x}^{L, l}_i \in \mathbb{R}^{d_l} )_{1 \leq i \leq n_l}$. At the coarsest level $L$, we set $\mathbb{X}^{L,L} = \mathbb{X}^{L}$. At level $l$, we unpool $\mathbb{X}^{L, l+1}$ to level $l$. This is done by unpooling to the vertex level, and employing max-pooling to all vertices of $(P^l_i)$, leading to features $\operatorname{Unpool}_{l+1,l}(\mathbb{X}^{L, l+1})$. We concatenate $\operatorname{Unpool}_{l+1,l}(\mathbb{X}^{L, l+1})$ and $\mathbb{X}^{l}$. This concatenated feature may be non-smooth on the surface along patch boundaries in hierarchy level $l+1$. To remedy this, we perform one FeaStConv convolution to get $\mathbb{X}^{L, l}$. The same computations yield $\mathbb{Y}^{L, l} = (\mathbb{y}^{L, l}_i \in \mathbb{R}^{d_l} )_{1 \leq i \leq m_l}$.

The association matrices between patches of $\mathcal{X}$ and $\mathcal{Y}$ at hierarchy level $l$ are computed using cosine similarity as a distance measure in feature space, similar to previous works \cite{eisenberger2021neuromorph, lang2021dpc}. We then employ softmax to get normalized similarity scores:

\noindent\begin{minipage}{.5\linewidth}
{\small 
\begin{equation}
\label{pi_x_y}
	{(\Pi^l_{\mathcal{X} \rightarrow \mathcal{Y}})}_{ij}
	:=
	\frac
	{\exp(s^l_{ij})}
	{\sum_{k=1}^{m_l}\exp(s^l_{ik})}
\end{equation}
}
\end{minipage}%
\begin{minipage}{.5\linewidth}
{\small
\begin{equation}
\label{pi_y_x}
	{(\Pi^l_{\mathcal{Y} \rightarrow \mathcal{X}})}_{ij}
	:=
	\frac
	{\exp(s^l_{ji})}
	{\sum_{k=1}^{n_l}\exp(s^l_{ki})}
\end{equation}
}
\end{minipage}

with $s^l_{ij} := \left(\langle{\mathbb{x}}^{L,l}_i,{\mathbb{y}}^{L,l}_j\rangle_2\right) /	\left(\|{\mathbb{x}}^{L,l}_i\|_2\|{\mathbb{y}}^{L,l}_j\|_2\right)$.

\subsection{Deformation Network}
\label{sec:deformation_network}
The deformation network deforms $\mathcal{X}$ to $\mathcal{Y}$ and $\mathcal{Y}$ to $\mathcal{X}$ while respecting association matrices $\Pi^l_{\mathcal{X} \rightarrow \mathcal{Y}}$ and $\Pi^l_{\mathcal{Y} \rightarrow \mathcal{X}}$, respectively, for every level $l$ of the hierarchy.

\textbf{Deformation Model}
We use a patch-based deformation model that represents a shape deformation as a collection of patch-wise near-rigid deformations~\cite{cagniart}. It allows to represent complex deformations while drastically reducing the deformation parameters. We summarize how to deform hierarchy level $l$ and omit subscript $l$ to simplify notation. Given $\mathcal{X}$  with its decomposition into surface patches  $(P_i)_{1 \leq i \leq n}$ along with their centers $\mathcal{C} = (c_i \in \mathbb{R}^3)_{1 \leq i \leq n}$, each patch $P_i$ is associated with a rotation matrix $R_i \in \mathbb{R}^{3 \times 3}$ and a new center position $u_i \in \mathbb{R}^3$. If we denote by $x_0(v) \in \mathbb{R}^3$ the original position of vertex $v$ in $\mathcal{X}$, the rigid deformation of a vertex $v$ according to $P_i$ can be written as $x_i(v) = R_i(x_0(v) - c_i) + u_i \in \mathbb{R}^3$. The rigid deformations of all patches $(P_i)_{1 \leq i \leq n}$ are blended at the vertex level using weighting functions $\alpha_i(v) \in \mathbb{R}$ defined as a Gaussian of the geodesic distance\footnote{In the original paper~\cite{cagniart}, the Euclidean distance is used to define the blending functions. We use the geodesic distance instead to account for points that might be extrinsically close to and intrinsically far from $c_i$.} of $v$ to $c_i$.

To ensure consistent deformations between neighboring patches $P_i$ and $P_j \in \mathcal{N}(P_i)$, transformations are constrained by minimizing
{\small
\begin{equation}
    \label{rig_loss}
    l_{rig}(\mathcal{X}) = \sum_{(P_i)_{1 \leq i \leq n}}{\sum_{P_j \in \mathcal{N}(P_i)}{\sum_{v \in P_i \bigcup P_j}{E^{ij}_v}}} \mbox{ with}
\end{equation}
\begin{equation}
    \label{rig_loss_term}
        E^{ij}_{v\in P_i \bigcup P_j} =  (\alpha_i(v) + \alpha_j(v)) \times \| x_i(v) - x_j(v) \|^2_2.
\end{equation}}

\textbf{Architecture}
Our network deforms both $\mathcal{X}$ to $\mathcal{Y}$ and $\mathcal{Y}$ to $\mathcal{X}$ symmetrically. We detail $\mathcal{X}$ to $\mathcal{Y}$ in what follows. At level $l$ of the hierarchy, we know matrix $\Pi^l_{\mathcal{X} \rightarrow \mathcal{Y}}$ from the association network. The deformation network consists of a feature extractor and a deformation decoder. 
The feature extractor is a convolutional neural network identical to the one used in the association network. It extracts patch-wise features for $\mathcal{X}$ and $\mathcal{Y}$ at level $l$ called $\Tilde{\mathbb{X}}^{l} = (\Tilde{\mathbb{x}}^l_i \in \mathbb{R}^{d_l})_{1 \leq i \leq n_l}$ and $\Tilde{\mathbb{Y}}^{l} = (\Tilde{\mathbb{y}}^l_i \in \mathbb{R}^{d_l})_{1 \leq i \leq m_l}$. We choose to decouple the features used for association and deformation to allow the networks to learn optimal features independently for each task.

The deformation decoder consists of a graph convolutional network followed by an MLP. It outputs rotation parameters $(R_i \in \mathbb{R}^6)_{1 \leq i \leq n_l}$ and new center positions $(u_i\in \mathbb{R}^3)_{1 \leq i \leq n_l}$ for every patch of level $l$ of $\mathcal{X}$. It takes as input patch centers $C_l^{\mathcal{X}} \in \mathbb{R}^{n_l \times 3}$, patch-wise features $\Tilde{\mathbb{X}}^{l}\in \mathbb{R}^{n_l \times d_l}$, patch centers of $\mathcal{X}$ projected in $\mathcal{Y}$ using the association matrix $\Pi^l_{\mathcal{X} \rightarrow \mathcal{Y}} C_l^{\mathcal{Y}} \in \mathbb{R}^{n_l \times 3}$, which represent spatial targets for the patch centers, and patch-wise features of $\mathcal{X}$ projected in patch-wise features of $\mathcal{Y}$ using the association matrix $\Pi^l_{\mathcal{X} \rightarrow \mathcal{Y}} \Tilde{\mathbb{Y}}^{l}\in \mathbb{R}^{n_l \times d_l}$.
We use the 6D representation of rotation matrices introduced in~\cite{6D_rotation} to allow for efficient learning. Applying the resulting transformations leads to the deformed shape $\mathcal{X}_l$.

\subsection{Learning Criteria}
We train our model using five self-supervised criteria at each hierarchy level $l$
{\small
\begin{equation}
    \label{final_loss}
    l_{network} = \sum_{l=0}^{L}{(\lambda^{l}_{\text{g}} l_{\text{geodesic}}^l +
        \lambda^{l}_{\text{c}} l_{\text{cycle}}^l + 
        \lambda^{l}_{\text{r}} l_{\text{rec}}^l 
    +\lambda^{l}_{\text{m}} l_{\text{match}}^l +
        \lambda^{l}_{\text{ri}} l_{\text{rigidity}}^l)},
\end{equation}}
where the individual loss terms for each level $l_{\text{geodesic}}^l, l_{\text{cycle}}^l, l_{\text{rec}}^l, l_{\text{match}}^l, l_{\text{rigidity}}^l$ are weighted by corresponding weights $\lambda^{l}_{\text{g}}, \lambda^{l}_{\text{c}}, \lambda^{l}_{\text{r}}, \lambda^{l}_{\text{m}}, \lambda^{l}_{\text{ri}}$ and detailed below.

\textbf{Geodesic Distance Distortion Criterion}
The first criterion favours maps that preserve geodesic distances, and is commonly used for shape matching~\eg~\cite{halimi2019unsupervised, eisenberger2021neuromorph}. We implement this by minimizing
$l_{\text{geodesic}}^l =  \|\Pi^l_{\mathcal{X} \rightarrow \mathcal{Y}}  D^l_{\mathcal{Y}} {\Pi^l_{\mathcal{X} \rightarrow \mathcal{Y}}}^T - D^l_{\mathcal{X}}\|^2_2 
    +  \|\Pi^l_{\mathcal{Y} \rightarrow \mathcal{X}}  D^l_{\mathcal{X}} {\Pi^l_{\mathcal{Y} \rightarrow \mathcal{X}}}^T$ $-D^l_{\mathcal{Y}}\|^2_2,$
where $D^l_{\mathcal{X}}$ and $D^l_{\mathcal{Y}}$ are the geodesic distance matrices of $\mathcal{X}$ and $\mathcal{Y}$, restricted to the patch centers of level $l$. This criterion measures the geodesic distortion of maps $\Pi^l_{\mathcal{X} \rightarrow \mathcal{Y}}$ and $\Pi^l_{\mathcal{Y} \rightarrow \mathcal{X}}$. We only use this loss on coarse levels because it is computationally prohibitive for finer levels.

\textbf{Cycle Consistency Criterion}
This criterion asks $\Pi^l_{\mathcal{X} \rightarrow \mathcal{Y}}$ and $\Pi^l_{\mathcal{Y} \rightarrow \mathcal{X}}$ to be cycle consistent, i.e.~every point going through a length two cycle is mapped back to itself. It is implemented by minimizing
$    l_{\text{cycle}}^l = \|\Pi^l_{\mathcal{X} \rightarrow \mathcal{Y}} (\Pi^l_{\mathcal{Y} \rightarrow \mathcal{X}} C^{\mathcal{X}}_{l}) - C^{\mathcal{X}}_{l}  \|^2_2 
    +  \|\Pi^l_{\mathcal{Y} \rightarrow \mathcal{X}} (\Pi^l_{\mathcal{X} \rightarrow \mathcal{Y}} C^{\mathcal{Y}}_{l}) - C^{\mathcal{Y}}_{l}  \|^2_2.$
 
\textbf{Self Reconstruction Criterion}
The third criterion aims to identify each patch in feature space, to avoid many patches being mapped to the same patch. This criterion helps handling intrinsically symmetric shapes, where different patches share the same intrinsic geometry. It is implemented by minimizing
$    l_{\text{rec}}^l =  \|\Pi^l_{\mathcal{X} \rightarrow \mathcal{X}} C^{\mathcal{X}}_{l}   - C^{\mathcal{X}}_{l}  \|^2_2 
    +  \|\Pi^l_{\mathcal{Y} \rightarrow \mathcal{Y}} C^{\mathcal{Y}}_{l}   - C^{\mathcal{Y}}_{l}  \|^2_2,$
where $\Pi^l_{\mathcal{X} \rightarrow \mathcal{X}}$ and $\Pi^l_{\mathcal{Y} \rightarrow \mathcal{Y}}$ are self association matrices computed similar to Eq.~\ref{pi_x_y} and \ref{pi_y_x}. 

\textbf{Matching Criterion}
The matching criterion aims to deform one shape to the other using the computed associations. It serves two roles. First, ensuring that the deformation realised by the deformation network matches the correspondences computed by the association network. Second, allowing for extrinsic geometric control on the correspondences. It is implemented by minimizing
$    l_{\text{match}}^l =  \|C^{\mathcal{X}_l}_{l} - \Pi^l_{\mathcal{X} \rightarrow \mathcal{Y}} C^{\mathcal{Y}}_{l}\|^2_2 
    + \|C^{\mathcal{Y}_l}_{l} - \Pi^l_{\mathcal{Y} \rightarrow \mathcal{X}} C^{\mathcal{X}}_{l}\|^2_2,$
where $C^{\mathcal{X}_l}_{l}$ and $C^{\mathcal{Y}_l}_{l}$  are the deformed cluster centers of $\mathcal{X}_l$ and $\mathcal{Y}_l$ respectively. 

\textbf{Rigidity Criterion}
The rigidity criterion aims for consistent transformations between neighboring patches at all hierarchy levels as
$    l_{\text{rigidity}}^l = l_{rig}(\mathcal{X}^l) + l_{rig}(\mathcal{Y}^l),$
with $l_{rig}$ defined in Eq.~\ref{rig_loss}. The matching and rigidity criteria encourage the network to explain the similarity in feature space on the deformed shapes in 3D at every level of the patch hierarchy. At every level of the hierarchy, the deformation for every vertex on the shapes is evaluated even though the matching criterion  evaluates the deformation result only on the patch centers of the deformed shapes. This means that even if the association matrices are sparse at coarse levels they consider the full correspondence between the two shapes. This acts as a strong inductive bias in our network, which promotes spatially continuous maps.

\textbf{Fine Tuning} Our full network is trained in an end-to-end fashion. After training, the network produces good quality matchings in a single forward pass. We improve these matchings by optimizing the network for each pair $\mathcal{X}$ and $\mathcal{Y}$ only, with the same architecture and hyper-parameters. This is possible as all of our losses are unsupervised and do not require any ground truth information. We refer to this specialisation as fine tuning. We extract final dense correspondences by applying argmax on vertex associations $\Pi^0_{\mathcal{X} \rightarrow \mathcal{Y}}$ and $\Pi^0_{\mathcal{Y} \rightarrow \mathcal{X}}$.

\section{Experiments}
\label{sec:experiments}

This section presents results of our method, including comparative evaluations, for human and animal shapes. For human shapes, we provide an extensive evaluation protocol to demonstrate that our method generalizes well to new body shapes and poses, sampling densities, topological noise, and raw acquisition data. In all of our experiments we fix the number of patches to $50$, $200$ and $800$. 
More details are provided in the appendices; Appendix~\ref{sec:implem_details} provides implementation details, Appendix~\ref{sec:data} provides details on data processing, Appendix~\ref{sec:ablation_study} provides an ablation study proving the benefits of the main parts of our method i.e. hierarchy in feature space, use of a deformation model and fine tuning, and Appendix~\ref{sec:additional_res} provides additional quantitative and qualitative comparisons and qualitative deformation results.

\textbf{Evaluation metrics} We consider two metrics that compare the ground truth mapping $\Pi^{gt}_{. \rightarrow .}$ with an estimated mapping $\Pi_{. \rightarrow .}$. The  mean geodesic error (MGE):
{\small
\begin{equation}
\label{mge_error}
	\operatorname{MGE}(\Pi_{\mathcal{X} \rightarrow \mathcal{Y}})
	:=
	\frac{1}{|\mathcal{X}|}
	\sum_{x \in \mathcal{X}}{d_{\mathcal{Y}}(\Pi_{\mathcal{X} \rightarrow \mathcal{Y}}(x), \Pi^{gt}_{\mathcal{X} \rightarrow \mathcal{Y}}(x))},
\end{equation}}
and the cycle geodesic error (CycleGE):
{\small
\begin{equation}
\label{cycle_error}
		\operatorname{CycleGE}(\Pi_{\mathcal{X} \rightarrow \mathcal{Y}}):=     	\frac{1}{|\mathcal{X}|^2}\sum_{\substack{x_1, x_2 \\ \in \mathcal{X} \times \mathcal{X} }}{|1-\frac{d_{\mathcal{X}}(x_1, x_2)}{d_{\mathcal{X}}(\Tilde{x_1}, \Tilde{x_2})}|}
		\textit{ with } \Tilde{x_{i}} :=  \Pi_{\mathcal{Y} \rightarrow \mathcal{X}}  (\Pi_{\mathcal{X} \rightarrow \mathcal{Y}}(x_{i})),
\end{equation}}
where $d_{\mathcal{X}}$, $d_{\mathcal{Y}}$ are the geodesic distances on meshes $\mathcal{X}$ and $\mathcal{Y}$, respectively, normalised by the square root of their area. These metrics allow to evaluate the distance between the true and estimated correspondences of a point $x$ on $\mathcal{Y}$, and the metric distortion induced by matching a pair of points using a length two cycle.

\textbf{Competing methods}  We compare our results against four state-of-the-art methods. FM based AttentiveFMaps~\cite{li2022attentivefmaps}. Spatial mesh matching method Neuromorph~\cite{eisenberger2021neuromorph} that guides the correspondences search using an interpolation while we search for a deformation. Deep Shells~\cite{eisenberger2020deep} an FM method that combines a spectral alignment with a spatial alignment. Point cloud matching method STS~\cite{efroni2022spectral} that uses an FM pipeline to extract global geometric properties, while we leverage spatial continuity at multiple scales. All methods that support supervised and unsupervised training were trained in the unsupervised regime.

\subsection{Evaluation on human data}

\hspace{\parindent} \textbf{Data} We train our method on clean pre-processed data, and test it both on (possibly degraded) pre-processed data and raw 3D acquisitions. The pre-processed data we use is the extended FAUST dataset~\cite{basset2021neural}, which extends FAUST~\cite{Bogo:CVPR:2014} with synthetically generated body shapes and poses. Extended FAUST contains 561 naked human meshes with 21 body shapes in roughly 27 poses. We select $451$ meshes for training and $110$ meshes for testing, where the test set contains body shapes and poses not used for training. We re-mesh each mesh independently and uniformly to about $5k$ vertices using~\cite{yan2009isotropic} to remove vertex density as a discriminative feature. We design test sets to evaluate different types of generalization.

\textbf{Generalization to unseen body shape and pose} The first test set considers pairs of models of the extended FAUST test set that differ in body shape and pose. It allows to evaluate how well methods generalize to body shapes and poses not observed during training.

\textbf{Robustness to sampling density} To evaluate robustness to different sampling strategies, we consider two test sets using resampled versions of the extended FAUST test set. The first one resamples the test set uniformly to contain $\approx 15k$ vertices. The second one resamples the test set with curvature adapted triangles~\cite{nivoliers2015anisotropic} to contain $\approx 5k$ vertices. 

\textbf{Generalization to different topology} Our method is particularly robust to changes in topology. To demonstrate this, we design a test set by altering the topology of the extended FAUST test set. Extended FAUST contains many shapes with near-contacts between different body parts, and we detect these parts and replace them by gluing the parts in contact. In practice, we achieve this by detecting self-intersections, deleting mesh parts located inside the mesh, and by filling the resulting holes using Poisson surface reconstruction~\cite{kazhdan2006poisson}. The final reconstructed surface contains $\approx 6k$ vertices.

\textbf{Generalization to raw 3D acquisitions} Our goal is to have a method that can be applied to raw scan data without pre-processing. Hence, we test our method on two test sets of raw acquisition data. All these data are plagued by acquisition noise, which alters the geometry or topology. The first one contains $17$ scans ($136$ pairs) of minimally dressed humans of CHUM~\cite{marsot-3dv-22} of two female subjects and one male subject, all in different poses. Each mesh contains about $15k$ vertices. The second one contains $12$ scans ($66$ pairs) of humans in everyday clothing of two female subjects and two male subjects of the dataset 4DHumanOutfit \cite{armando20234dhumanoutfit}, all in different poses. These scans were captured in a multi-view acquisition platform with $68$ synchronized cameras.

\textbf{Comparison to state-of-the-art}
We first provide a quantitative evaluation with Neuromorph, AttentiveFMaps and Deep Shells, where all methods were trained on the extended FAUST training set and tested on the different test sets. Tab.~\ref{tab:merged_xlf} shows the results. We could not include STS in this comparison, the code for pre-processing new data is not provided, making training or testing on our data difficult. 

\begin{table}[ht]
\centering
\resizebox{1.0\columnwidth}{!}{\begin{tabular}{|l||c|c|c|c|c|c|c|c||c|c|c|c|}
    \hline
    \multirow{3}{*}{Method} & \multicolumn{8}{|c||}{Pre-processed data} &  \multicolumn{4}{|c|}{Raw 3D acquisitions} \\\cline{2-13}
        & \multicolumn{2}{|c|}{Body shape and pose} & \multicolumn{4}{|c|}{Sampling density} & \multicolumn{2}{|c||}{Topology} & \multicolumn{2}{|c|}{Naked} & \multicolumn{2}{|c|}{Clothed} \\
        \cline{4-7}
        & \multicolumn{2}{|c|}{ } & \multicolumn{2}{|c|}{Uniform $15K$ vert.} & \multicolumn{2}{|c|}{Non-uniform $5K$ vert.} & \multicolumn{2}{|c||}{ } & \multicolumn{2}{|c|}{ } & \multicolumn{2}{|c|}{ }\\
        \hline
        \hline
    Deep Shells & 13.39 & 0.0694 & 9.11 & 0.0564 & 17.10 & 0.1409 & 20.48 & 0.1608 & 16.40 & 0.0775 & 26.49 & 0.2275 \\ \hline
    Neuromorph & 11.61 & 0.1315 & 14.45 & 0.1527 & 37.87 & 0.5389 & \underline{11.92} & \underline{0.1231} & 13.83 & \underline{0.1407} & \underline{16.01} & \underline{0.1429} \\ \hline
    AttentiveFMaps &\textbf{ 2.65} & \textbf{0.0171} & \textbf{2.47} & \textbf{0.0160} & \textbf{2.82} & \textbf{0.0385} & 18.93 & 0.3426 & \underline{11.38} & 0.1608 & 49.19 & 0.7701 \\ \hline
    Ours & \underline{4.09} & \underline{0.0259} & \underline{4.15} & \underline{0.0304} & \underline{5.85} & \underline{0.0552} & \textbf{8.22} & \textbf{0.0441} & \textbf{3.74} & \textbf{0.0349} & \textbf{6.02} & \textbf{0.0463} \\ \hline
  \end{tabular}}
\caption{Comparison to state-of-the-art on human data. Each column shows MGE (Eq.~\ref{mge_error}) on left and CycleGE (Eq.~\ref{cycle_error}) on right. Best scores in bold, second best scores underlined.\label{tab:merged_xlf}}
\end{table}

For all test sets, our method outperforms Neuromorph and Deep Shells. Furthermore, our method performs on-par overall with AttentiveFMaps for pre-processed data. While AttentiveFMaps performs slightly better than our method when applied to clean pre-processed data (unseen body shape and pose and different sampling density columns in Tab.~\ref{tab:merged_xlf}), it degrades significantly in the presence of topology changes. We believe that this is due to the global nature of spectral shape decompositions that naturally favours clean data. Our method significantly outperforms AttentiveFMaps in this case as our method generalizes well to topological errors. Our approach builds on a more local strategy, though also considering global geometric shape properties. This results in rare failure cases with important symmetry ambiguities but provides a strongly increased robustness to noise present in real situations, with raw acquisition data for example.

For raw acquisition data, our method significantly outperforms all other methods, demonstrating its practical value. This is thanks to its robustness to geometric and topological noise which characterize raw 3D acquisitions.
Fig.~\ref{fig:qualitataive_eval} (top row) visualizes an example result of a raw 3D data acquisition in everyday clothing. The target is color coded and the colors are transferred using the correspondences computed by the different methods. Our method is correct both globally and locally, while competing methods tend to fail in this case. Our method is designed for complete shape matching, thus we only report results on shape acquisitions that are complete. Our method fails locally on partial shapes.

To compare to STS, we run our method on their training and test sets of FAUST~\cite{Bogo:CVPR:2014}, with their evaluation protocol considering point-to-point accuracy\footnote{This metric measures the proportion of points for which the geodesic error in Eq.~\ref{mge_error} is equal to zero.} and MGE\footnote{In this case the geodesic error is normalised by the geodesic diameter rather than the mesh surface.}. While STS reports $50.5\%$ and $9.5$, our method's results are $22.2\%$ and $1.6$ for point-to-point accuracy and MGE, respectively. Hence, our method outperforms STS.

\subsection{Evaluation on animal data}
\begin{wraptable}{R}{0.4\textwidth}
\centering
\resizebox{0.9\linewidth}{!}{
\begin{tabular}{ |l||c|c| }
 \hline
 Method & \thead{Mean \\ Geodesic \\ Error} & \thead{Cycle \\ Geodesic \\ Error} \\
 
\hline
\hline

{Deep Shells}      & 21.40  & 0.1984 \\
\hline
{Neuromorph}   & 12.34 & 0.1695 \\ \hline

{AttentiveFMaps}           & \textbf{4.40} & \underline{0.0601}    \\
 \hline

{Ours}               & \underline{5.23} &  \textbf{0.0562}    \\
\hline
\end{tabular}}
\caption{Comparison on animal shapes w.r.t.~MGE (Eq.~\ref{mge_error}) on left and CycleGE (Eq.~\ref{cycle_error}) on right. Best score in bold, second-best underlined.}
\label{tab:mge_smal}
\end{wraptable}

\textbf{Data} We consider pre-processed animal models sampled from the SMAL animal model~\cite{zuffi20173d}. We consider $49$ animal meshes re-meshed independently to about $5k$ vertices. We use $32$ meshes for training and $17$ ($136$ pairs) for testing.

\textbf{Comparison to state-of-the-art}
Tab.~\ref{tab:mge_smal} shows the results for Deep Shells, Neuromorph, AttentiveFMaps and our method. Our method's performance is better than Deep Shells and Neuromorph, and slightly lower than AttentiveFMaps. Fig.~\ref{fig:qualitataive_eval} (bottom row) visualizes an example result. Our method is visually on par with AttentiveFMaps. This shows that our method successfully applies to data beyond human models.

\section{Conclusions}
\label{sec:conclusions}

We have presented a data-driven unsupervised approach to solve for non-rigid shape matching. Our approach uses an association network that embeds the shapes in a feature space where hierarchical maps are extracted. It then constrains these maps by extracting the induced alignment in 3D using a deformation network that fits a piece-wise near-rigid deformation model. Our method retains the robustness of spatial methods while enforcing global geometric constraints on the associations, as with spectral methods. We demonstrate experimentally that our approach performs on-par with state-of-the-art for pre-processed data and significantly outperforms existing methods when applied directly to the raw output of multi-view 3D reconstructions.

\section{Acknowledgements}
\paragraph{}
We thank Abdelmouttaleb Dakri for providing us with SMPL fittings for our experiments. This work was funded by the ANR project Human4D (ANR-19-CE23-0020). 

\bibliography{main}

\clearpage
\appendix
This supplementary material provides implementation details of our method in Appendix~\ref{sec:implem_details}, more details on the datasets used in Appendix~\ref{sec:data}, ablation studies in Appendix~\ref{sec:ablation_study}, and  additional quantitative and qualitative results in Appendix~\ref{sec:additional_res}.


\section{Implementation Details}
\label{sec:implem_details}

\subsection{Patch Extraction}
\label{sec:surface_patches}
To compute multi-resolution surface patches, we consider a  greedy approach based on the furthest point sampling strategy inspired by \cite{peyre2006geodesic}. Starting from a randomly selected vertex $x_1$, we compute the geodesic distance map $U_{x_1}$ to all other vertices on the mesh. Then, given that we have a set of vertices $S_n = \{x_1,..,x_n\}$, and their distance map $U_n$, we select the new vertex $x_{n+1}$ to be the furthest vertex from $S_n$. We compute $U_{x_{n+1}}$, the distance map from $x_{n+1}$ and update $U_{n+1} = \operatorname{min}(U_n, U_{x_{n+1}})$ and add $x_{n+1}$ to $S_n$ to get $S_{n+1}$. We stop the algorithm when a target number of points is reached. To get multiple patch resolutions we stop the algorithm at increasing numbers of target points. 

We select $L+1$ patch resolutions where patches at level  $0$  are restricted to vertices and level $L$ represents the coarsest patch level. For each hierarchical level, the selected samples are used as patch centers $\mathcal{C}_l = (c^l_i \in \mathbb{R}^3)_{1 \leq i \leq n_l}$ and their corresponding Voronoi cells on the mesh as patches $(P^l_i)_{1 \leq i \leq n_l}$ for $l=0,\ldots,L$. 

\begin{figure}[htb]
\centering
  \includegraphics[width=0.5\textwidth]{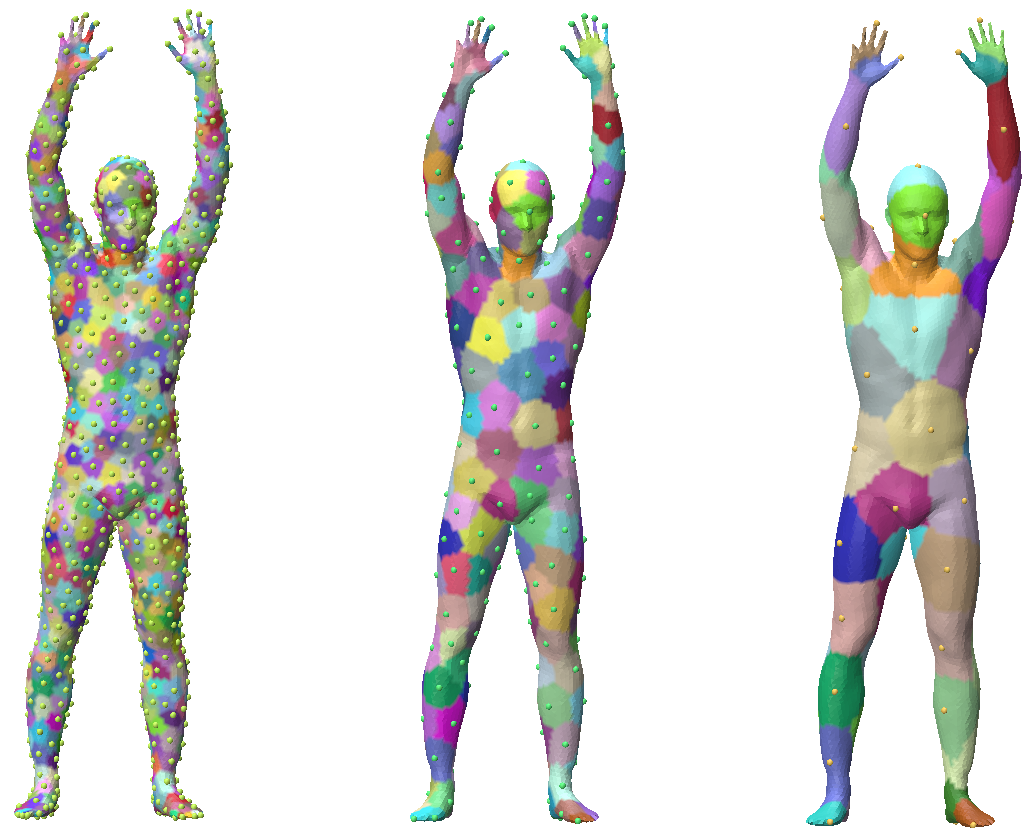}
  \caption{Example of surface patches. From left to right : $800$, $200$ and $50$ patches along with their centers on a mesh with $\approx 13k$ vertices.}
  \label{fig:path_hierarchy}
\end{figure}

In all of our experiments, we extract $4$ patch resolutions : all the mesh vertices, $800$, $200$ and $50$ patches. Figure~\ref{fig:path_hierarchy} shows an example of these surface patches.

\subsection{Architecture Detail}

\paragraph{Pooling and Unpooling} The extracted surface patches are not strictly hierarchical in the sense that  vertices of a patch at level $l$ can belong to multiple patches at coarser levels $l+1$. We use these surface patches for pooling and unpooling operations in our architectures.

To pool features from patches $(P^l_i)_{1 \leq i \leq n_l}$ of level $l$ to coarser patches $(P^{l+1}_i)_{1 \leq i \leq n_{l+1}}$ of level $l+1$ we proceed in two step:
\begin{enumerate}
    \item We first unpool to the vertex level (level $0$) such that each vertex is associated with the features of the patch it belongs to in level $l$.
    \item  We then employ max-pooling to go to patch level $l+1$.
\end{enumerate}  

To unpool features from patches $(P^{l+1}_i)_{1 \leq i \leq n_{l+1}}$ of level $l+1$ to finer patches $(P^l_i)_{1 \leq i \leq n_l}$ of level $l$ we proceed similarly. We first unpool to the vertex level (level $0$) such that each vertex is associated with the features of the patch it belongs to in level $l+1$. We then employ max-pooling to go to patch level $l$.

\begin{figure}[htb]
\centering
  \includegraphics[width=1.0\textwidth]{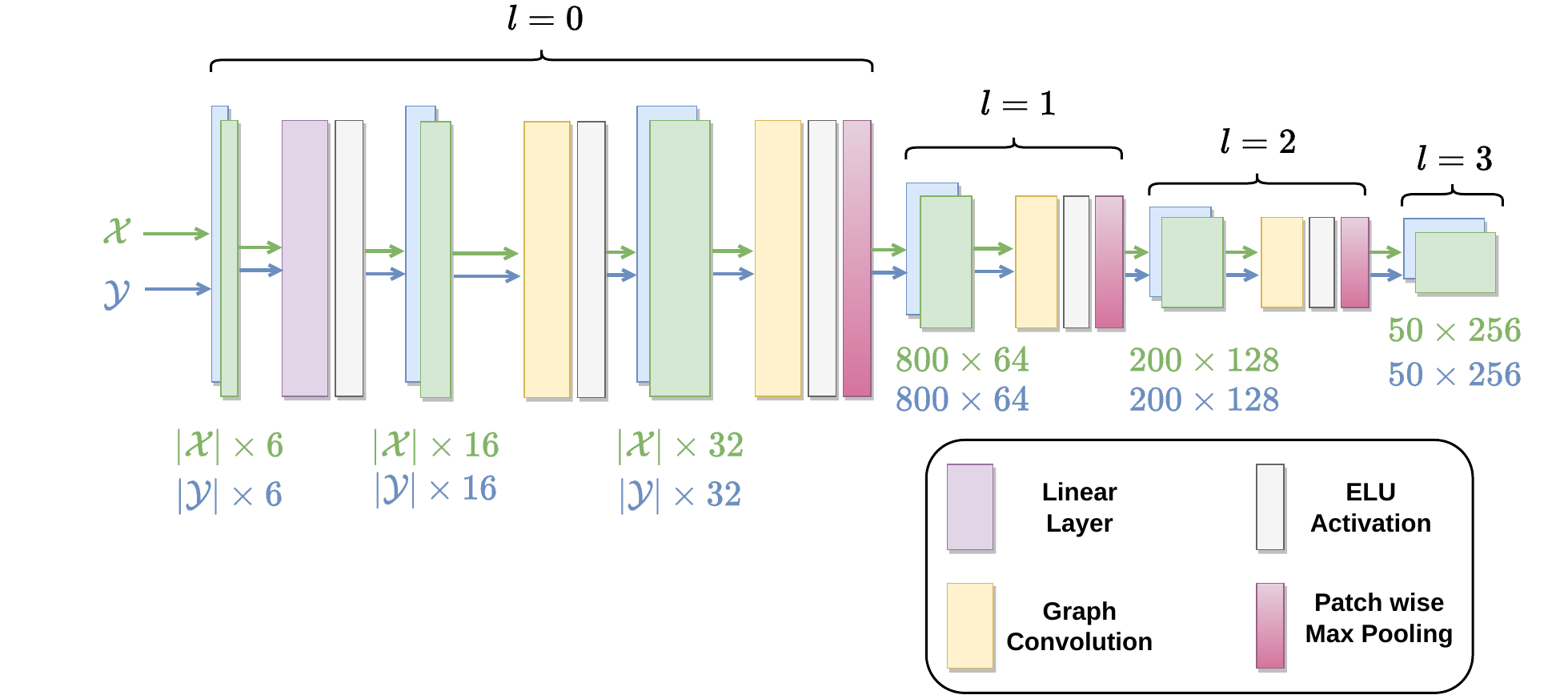}
  \caption{Feature extractor architecture. Inputs $\mathcal{X}$ and $\mathcal{Y}$ are decomposed into a hierarchy of $800$, $200$ and $50$ patches and fine-to-coarse features are extracted. Input features at the vertex level are 3D coordinates and normals.}
  \label{fig:feature_extractor}
\end{figure}

\paragraph{Feature Extractor} In both the association and the deformation networks, we use identical feature extractors based on  hierarchical graph convolutional network FeaStConv operators~\cite{verma2018feastnet}. In all cases, we fixed the number of attention heads of this operator to $9$. Figure~\ref{fig:feature_extractor} illustrates the feature extractor architecture.

\begin{figure}[htb]
\centering
  \includegraphics[width=1.0\textwidth]{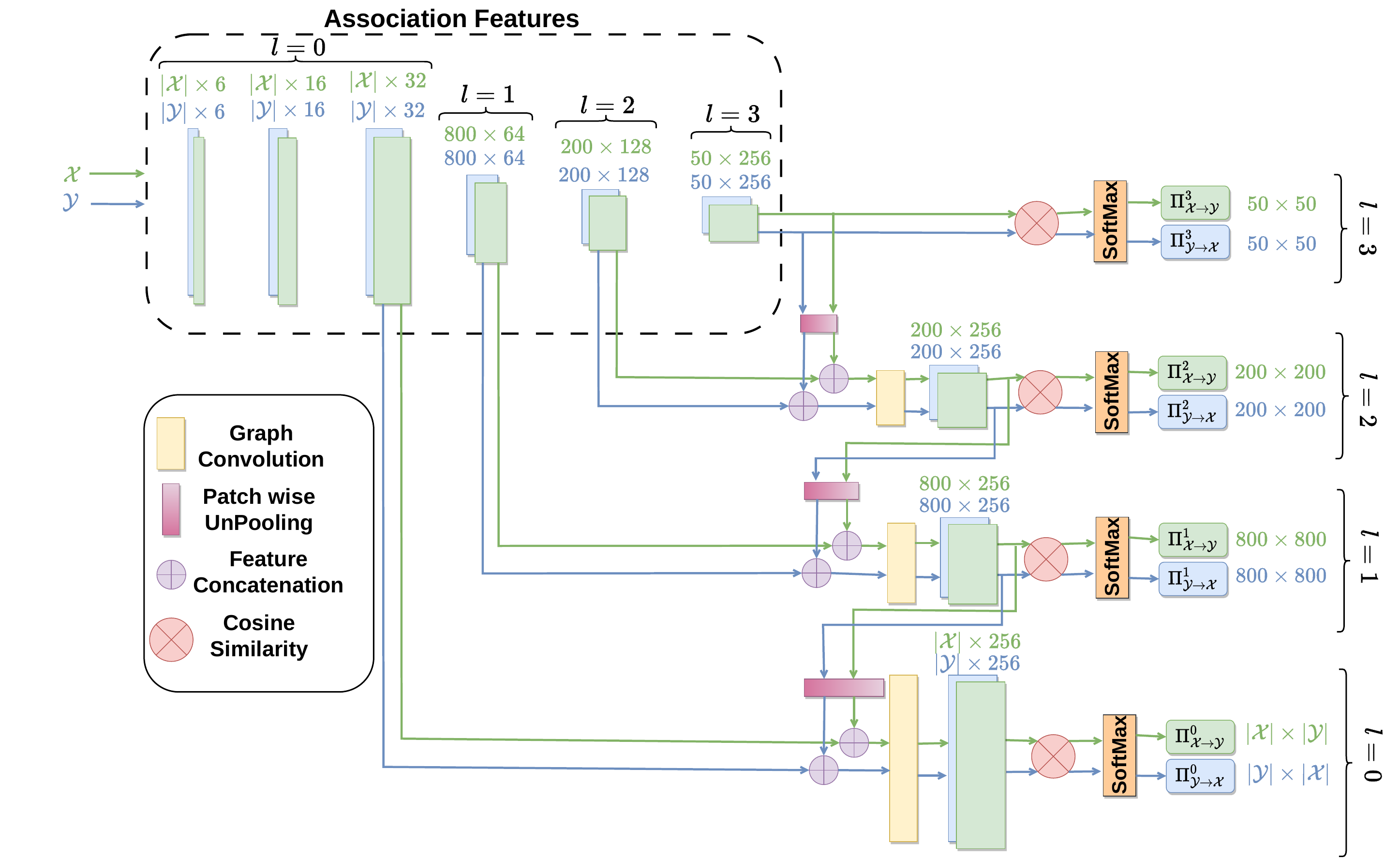}
  \caption{Architecture of the association network. Given meshes $\mathcal{X}$ and $\mathcal{Y}$ it outputs coarse-to-fine association maps at every level of the patch hierarchy, i.e. $\Pi^l_{\mathcal{X} \rightarrow \mathcal{Y}}$ and $\Pi^l_{\mathcal{Y} \rightarrow \mathcal{X}}$ for every level $l$.}
  \label{fig:association_network}
\end{figure}

\paragraph{Association Network}
  Figure~\ref{fig:association_network} details the architecture of the association network. We fix the temperature parameter of the softmax operator to $10^{-2}$.

\begin{figure}[htb]
\centering
  \includegraphics[width=1.0\textwidth]{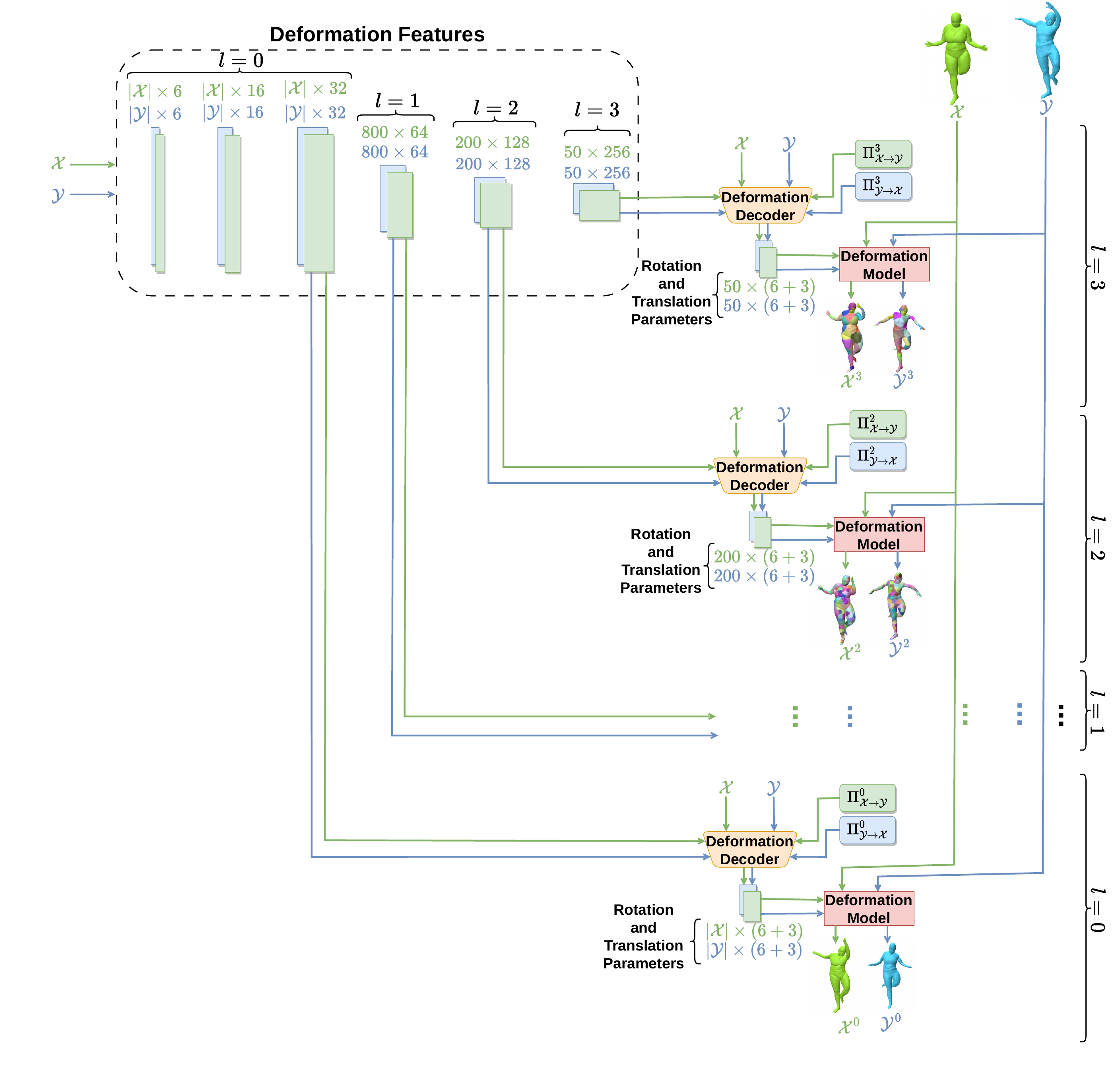}
  \caption{Architecture of the deformation network. Given  $\mathcal{X}$, $\mathcal{Y}$, $\Pi^l_{\mathcal{X} \rightarrow \mathcal{Y}}$, and $\Pi^l_{\mathcal{Y} \rightarrow \mathcal{X}}$ it estimates the 3D deformations $\mathcal{X}^{l}$ and $\mathcal{Y}^{l}$  of $\mathcal{X}$ and $\mathcal{Y}$ for every level $l$.}
  \label{fig:deformation_network}
\end{figure}

\begin{figure}[htb]
\centering
  \includegraphics[width=1.0\textwidth]{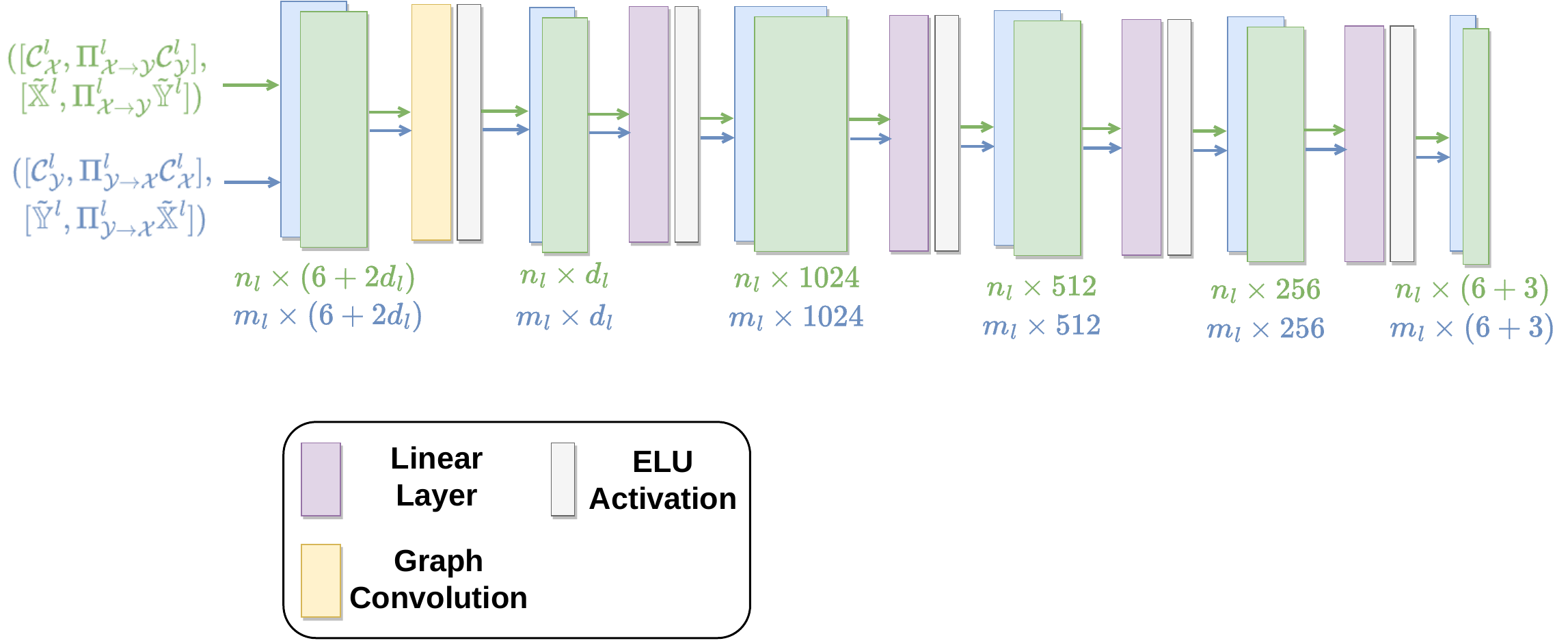}
  \caption{Architecture of the deformation decoder. It is composed of a graph convolution followed by an MLP.}
  \label{fig:deformation_decoder}
\end{figure}

\paragraph{Deformation Network}
Figure~\ref{fig:deformation_network} details the architecture of the deformation network. It uses a deformation decoder module that outputs per-patch rotation and translation parameters detailed in Figure~\ref{fig:deformation_decoder}.

\subsection{Training and Inference}
\paragraph{Loss Weights}
The final loss per level $l$ is the weighted combination of five self-supervised criteria, as given in Equation~\ref{final_loss} of the main paper. Table~\ref{tab:loss_weights} shows the weights used to train the network.
\begin{table}[ht]
\centering
\resizebox{0.5\columnwidth}{!}{
\begin{tabular}{ |c||c|c|c|c| }
 \hline
 \backslashbox{Weight}{Level}  & $l = 0$ & $l = 1$ & $l = 2$ & $l = 3$ \\
 \hline
 \hline
 $\lambda_g^l$ & 0.0 & 1.0 & 1.0 & 1.0 \\
 \hline
 $\lambda_c^l$ & 10.0 & 10.0 & 10.0 & 10.0 \\
 \hline
 $\lambda_r^l$ & 10.0 & 10.0 & 10.0 & 10.0 \\
 \hline 
 $\lambda_m^l$ & 2.0 & 2.0 & 2.0 & 2.0 \\
 \hline
 $\lambda_{ri}^l$ & 200.0 & 100.0 & 50.0 & 10.0 \\
 \hline
  
\end{tabular}}
\caption{Loss weights for every level.}
\label{tab:loss_weights}
\end{table}
The geodesic criterion was not used at the vertex level (weight fixed to $0$) as it involves matrix multiplications that are computationally prohibitive. The matching loss and the rigidity loss are somehow in opposition: The rigidity loss promotes isometric deformations whereas the matching loss promotes deformations that reflect the association matrix. Their weights are fixed so as to satisfy the association matrices while preserving the spatial continuity at the patch borders.

\paragraph{Training}
Our network is trained using Adam~\cite{kingma2014adam} with gradient clipping. The learning rate is fixed to $1\times10^{-3}$ for the first epoch, $5\times10^{-4}$ between the $2^{nd}$ and $10^{th}$ epochs and $2.5\times10^{-4}$ after the $10^{th}$ epoch. Our model takes $372$ epochs to train on the Extended FAUST training set. We select the model that achieves the smallest loss on a validation set.

\paragraph{Inference}
At test time we allow the network to specialise for each new shape pair to improve the matching. This is achieved by resuming the training of the selected model on a training set restricted to the two input shapes, with the same fixed architecture, optimization technique and hyper-parameters. We refer to this specialisation as fine tuning in the main paper. In all our experiments we fix the number of epochs for fine tuning to $50$.


\section{Data}
\label{sec:data}
We detail below how ground truth correspondences were obtained to evaluate our approach. 

\paragraph{Pre-processed Data}
For our experiments on pre-processed data, we used the extended FAUST dataset~\cite{basset2021neural} which is composed of meshes with the same template connectivity. In order to remove the connectivity consistency, which can strongly bias the matching, we remeshed all shapes individually and created the $3$ test sets mentioned in the main paper (sec 4.1). To get the ground truth correspondences for vertices on the remeshed shapes, we revert to the connectivity consistent meshes by searching, for each vertex  on a remeshed shape, the triangle on the original mesh, with minimal distance along the vertex normal direction. As a result of the connectivity changes, such distance can be large and we discard in the evaluation vertices for which this distance is higher than $2/10$ of the remeshed shape's mean edge length. This corresponds to $4.87\%$ of the vertices for uniform remeshings to $5k$ vertices, $1.2\%$ for uniform remeshing to $15k$ vertices and $1.01\%$ for curvature adapted remeshings to $5k$ vertices. A similar strategy is used  for meshes altered with topological noise. In this case, $7.69\%$ of the vertices are discarded in the evaluation.

\paragraph{Raw 3D Acquisition}
We also experimented our matching approach with raw 3D scans, \eg~\cite{marsot-3dv-22}. In this case, ground truth matchings were obtained by fitting the SMPL model~\cite{loper2015smpl} to the scans and considering closest vertices on the template in the normal direction.  Again here, vertices with distances to the template higher than $2$ times  the scan’s mean edge length are discarded in the evaluation. This corresponds in practice to $8.04\%$ for the naked raw acquisition dataset and $17.42\%$ for the clothed one.


\section{Ablation Studies}
\label{sec:ablation_study}
We present ablation studies that evaluate  the respective benefits of the main components of our approach, i.e.~the hierarchical modeling and the deformation model constraint. The impact of fine tuning is also given. To assess the hierarchical modeling, we use only two levels of hierarchy: the mesh vertices and $800$ patches. To assess the deformation model constraint, the network is restricted to the association network, losses involving the deformations, i.e.~matching and rigidity loss, are discarded.  
We trained the models on extended FAUST  and tested on both extended FAUST  and the raw 3D acquisition data with everyday clothing. Tab.~\ref{tab:ablation_table} shows the results and the number of epochs required to train each model. The complete model achieves the best results on both pre-processed and raw acquisition data. Note that the dimensionality reduction using the hierarchical modeling of associations is essential to the  unsupervised learning and allows for much faster training. Note also that the deformation model improves the quality of the matching and that the fine tuning improves the results across all models.

\begin{table}[ht]
\centering
\resizebox{0.8\columnwidth}{!}{
\begin{tabular}{ |c|c|c||c||c|c||c|c| }
 \hline
 \multirow{2}{*}{\thead{Hierarchical \\ Feature \\ Space}} &  \multirow{2}{*}{\thead{Deformation \\ Model}} &  \multirow{2}{*}{\thead{Fine \\ Tuning}} & 
 \multirow{2}{*}{\thead{Number of \\ Epochs \\ to Train}} & \multicolumn{2}{c||}{Pre-processed data} & \multicolumn{2}{c|}{Raw 3D acquisitions}  \\ \cline{5-8}
 &&&&\thead{Cycle \\ Geodesic \\ Error} & \thead{Mean \\ Geodesic \\ Error}& \thead{Cycle \\ Geodesic \\ Error} & \thead{Mean \\ Geodesic \\ Error} \\
 \hline
 \hline
  \cmark & \cmark & \cmark & 372& \textbf{0.0259} & \textbf{4.09}& \textbf{0.0463} & \textbf{6.02}  \\
 \hline
 \cmark & \cmark & \xmark & 372& 0.0352 & 4.99 &  0.0980 & 9.13  \\
 \hline
 \xmark & \cmark & \cmark & 803&  0.0653 & 14.01&  0.0556 & 9.41  \\
 \hline
  \xmark & \cmark & \xmark & 803& 0.0855 & 14.49& 0.0890 & 11.26 \\
 \hline
  \cmark & \xmark & \cmark & 311& 0.0260 & 4.24& 0.0519 & 6.44 \\
 \hline
  \cmark & \xmark & \xmark & 311& 0.0336 & 4.88 & 0.1247  & 10.77 \\
 \hline

\end{tabular}}
\caption{Ablation tests for the  hierarchy in feature space, the deformation model and the fine tuning.}
\label{tab:ablation_table}
\end{table}

To prove the added benefit of the Self-Reconstruction Criterion that ensures that each patch, on the shape itself, is identified to avoid many-to-one matches, we present an ablation where we both train and test the models on extended FAUST. Tab.~\ref{tab:ablation_rec} shows the results.

\begin{table}[ht]
\centering
\resizebox{0.4\columnwidth}{!}{
\begin{tabular}{|c|c||c|c| }
 \hline
 \thead{Self\\ Reconstruction\\Criterion} & \thead{Fine \\ Tuning} & \thead{Cycle \\ Geodesic \\ Error} & \thead{Mean \\ Geodesic \\ Error}  \\
 \hline
 \hline
 \cmark & \xmark & \textbf{0.0352} & \textbf{4.99}  \\
 \hline
 
  \xmark & \xmark & {0.0390} & {6.90} \\
 \hline

 \hline
\end{tabular}}
\caption{Ablation test of the Self-Reconstruction Criterion.}
\label{tab:ablation_rec}
\end{table}


\section{Additional Results}
\label{sec:additional_res}
\subsection{Additional Quantitative Evaluation}

\begin{figure}
\centering
\begin{tabular}{cc}
\includegraphics[width=4cm]{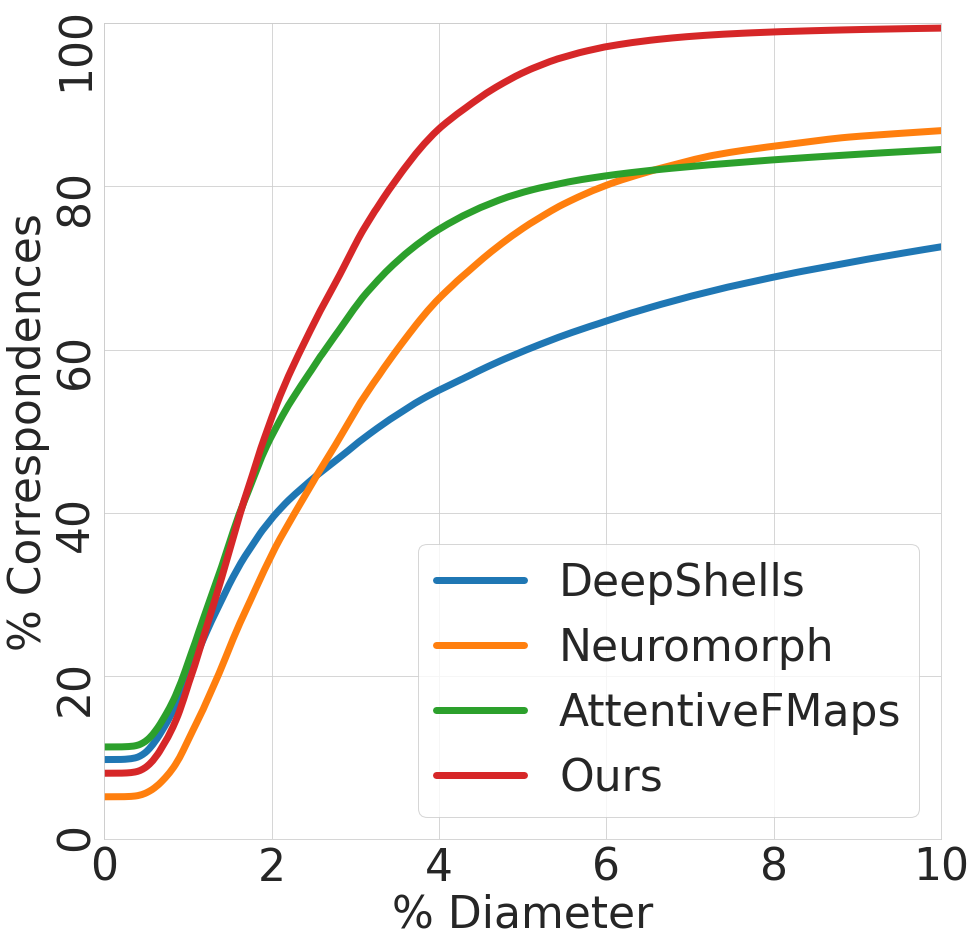} &\includegraphics[width=4cm]{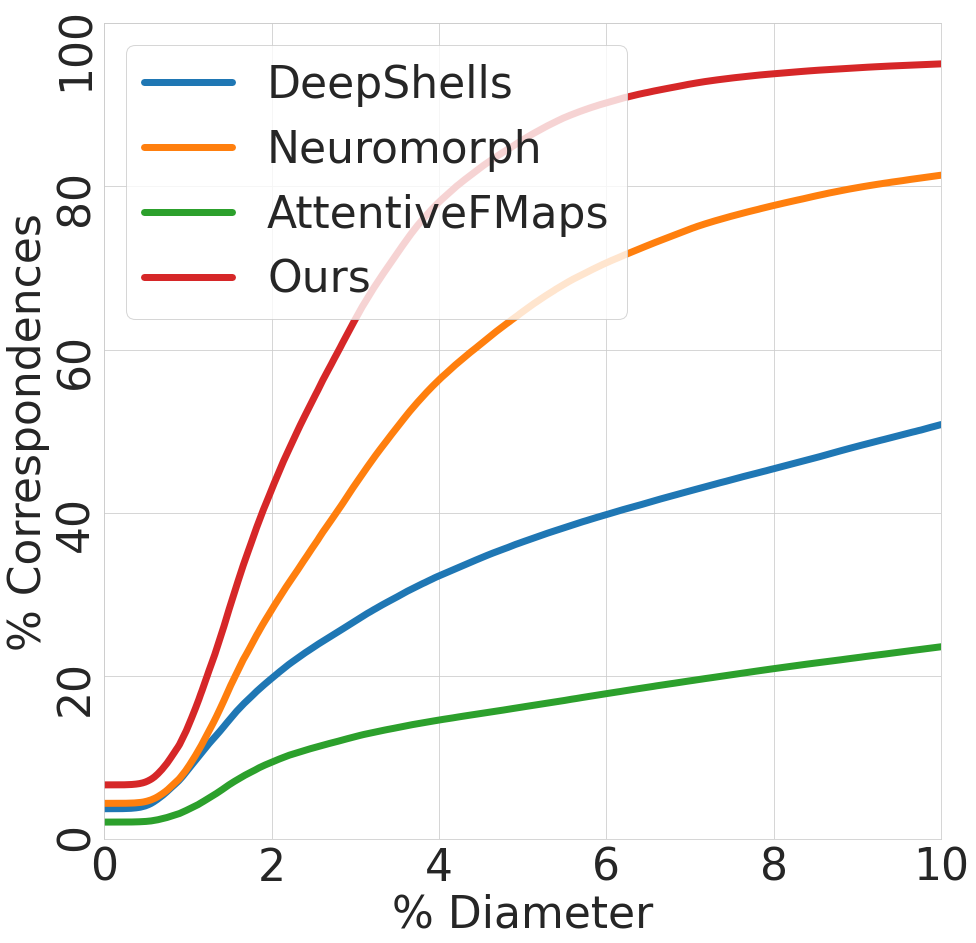}\\

(a) Naked raw acquisition data. &(b) Clothed raw acquisition data.
\end{tabular}

\caption{Comparison to state-of-the-art on raw acquisition data. Percentage of correct correspondences within a certain geodesic error tolerance radius.}
\label{fig:xlf_on_kinovis}
\end{figure}

Fig.~\ref{fig:xlf_on_kinovis} shows cumulative error plots giving the percentage of correct correspondences within a certain tolerance radius of geodesic error for the raw acquisitions in the naked regime in (a) and with everyday clothing in (b). On naked raw acquisitions, our method achieves $99\%$ of exact matches when we tolerate errors smaller then $8.1\%$ of the geodesic diameter, where the second best i.e.~Neuromorph achieves $85.1\%$. On clothed raw acquisitions, which is the most challenging test dataset, our method is more accurate both when considering details (i.e.~small errors) and global alignments (i.e.~large errors).

\subsection{Additional Qualitative Comparisons}

\begin{figure}[htb]
\centering
\includegraphics[width=0.9\textwidth]{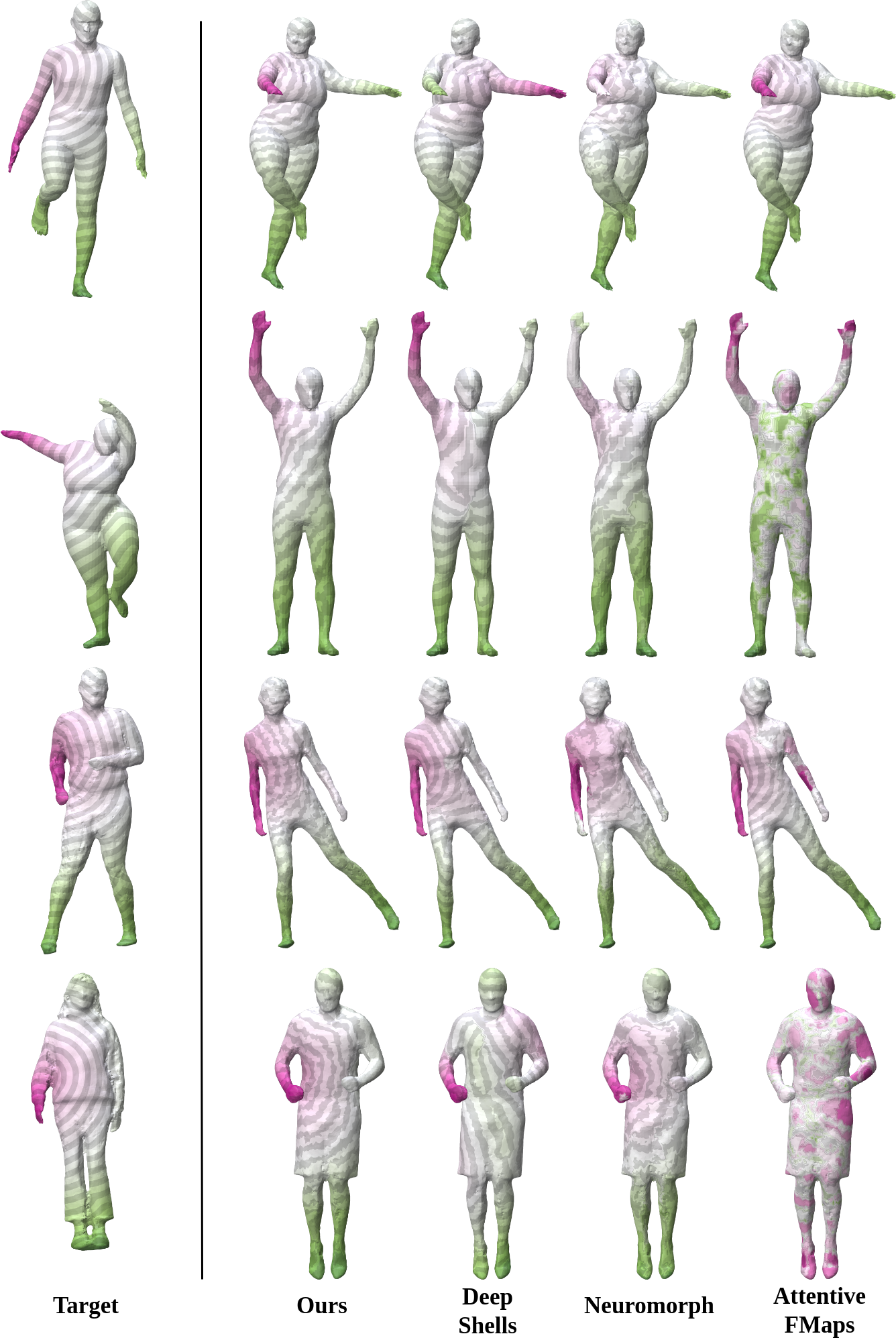}
\caption{Comparisons with Deep Shells \cite{eisenberger2020deep}, Neuromorph \cite{eisenberger2021neuromorph} and AttentiveFMaps \cite{li2022attentivefmaps} on pre-processed human meshes (first and second rows) and on raw human 3D scans (third and forth rows). Each point on the target mesh (left) is assigned a color, which is transferred to the source mesh (right) using the correspondences estimated by the different methods.}
\label{fig:additional_human}
\end{figure}

Fig.~\ref{fig:additional_human} shows qualitative comparisons on pre-processed meshes in the first row, pre-processed meshes with topological noise in the second row (the left heel is glued to the right calf and the left arm is glued to the head on the target shape), naked raw 3D scans in the third row and clothed raw 3D scans in the fourth row. The target is color coded and the colors are transferred using the correspondences as estimated by the different methods. Ours is both locally and globally accurate in all four cases and is robust to the presence of hairs, clothes and severe topological noises (see the last row). DeepShells makes global alignment errors (~\eg the arms are flipped in the first row, the belly area also in the other rows). Neuromorph suffers from local distortions and makes local errors (~\eg see the right hand in all rows). AttentiveFMaps is globally and locally accurate on  the pre-processed meshes in the first row but fails in the presence of topological noise which is ubiquitous in raw scans (~\eg see the full body in the second and fourth rows and the left arm area in the third row). 
\subsection{Qualitative Deformation Results}
\begin{figure}[htb]
\centering
\includegraphics[width=0.9\textwidth]{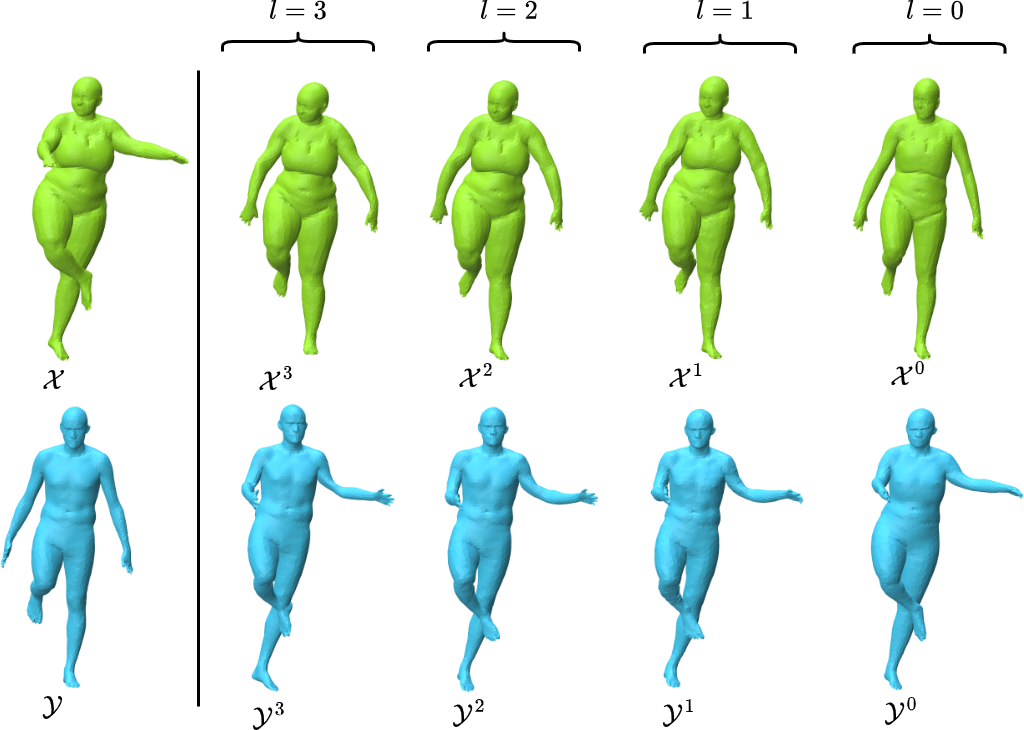}
\caption{Deformation examples output by our network for every level in the hierarchy. The top row shows the deformation results in the $\mathcal{X} \rightarrow \mathcal{Y}$ direction and the bottom row shows the deformation results in the $\mathcal{Y} \rightarrow \mathcal{X}$ direction.}
\label{fig:deformation_results}
\end{figure}

Fig.~\ref{fig:deformation_results} shows an example of deformed shapes output by our network at every hierarchical level. The top row shows the deformation in the $\mathcal{X} \rightarrow \mathcal{Y}$ direction and the bottom row shows the deformation in the $\mathcal{Y} \rightarrow \mathcal{X}$ direction. The deformed shapes at the finest level, i.e. the vertex level are close to the target deformation shapes ($\mathcal{X}^0 \approx \mathcal{Y}$ and $\mathcal{Y}^0 \approx \mathcal{X}$). In coarser levels, the deformation approximates the pose (global alignment) while in finer levels, where the rigidity constraint is weaker, it approximates the body shape (local alignment). This deformation is the induced alignment in 3D that guides the matching output of our method as shown in the top row of Fig.~\ref{fig:additional_human}.

\end{document}